\documentclass[runningheads]{llncs}

\usepackage{arydshln}   % For dashed and dotted lines in tables
\usepackage{tcolorbox}  % For text boxes
\usepackage{rotating}       % For rotating the table

\usepackage{colortbl}  % Required for cell coloring
\usepackage{xcolor}    % Required for text coloring

\usepackage{subcaption} % Add this to your preamble
\usepackage[labelfont=bf]{caption}
\usepackage{amsmath}
\usepackage{algorithm}
\usepackage{algpseudocode}
\usepackage{multirow}
\DeclareMathOperator*{\argmax}{arg\,max} 
\DeclareMathOperator{\loss}{\ell}
\DeclareMathOperator{\Loss}{\mathcal{L}}

\usepackage[T1]{fontenc}
\usepackage{amsfonts}
\usepackage{graphicx}
\usepackage{cite}
\usepackage{adjustbox}

\begin{document}
\title{Dynamical Label Augmentation and Calibration for Noisy Electronic Health Records}
\titlerunning{Dynamical Label Augmentation and Calibration for noisy EHR}
% \orcidID{0009-0003-5757-3715}

% comment out for blind review
\author{Yuhao Li\inst{1} \thanks{Corresponding author: yuhao.li4@unimelb.edu.au}
\and
Ling Luo\inst{1} \and
Uwe Aickelin\inst{1}}

\authorrunning{Li et al.}

% \authorrunning{Li et al.}
\institute{
School of Computing and Information Systems, University of Melbourne\\ %Melbourne, Parkville, VIC 3052, Australia\\
\email{\{yuhao.li4, ling.luo, uwe.aickelin\}@unimelb.edu.au} \\
}

\maketitle        % typeset the header of the contribution
\begin{abstract}
Medical research, particularly in predicting patient outcomes, heavily relies on medical time series data extracted from Electronic Health Records (EHR), which provide extensive information on patient histories. Despite rigorous examination, labeling errors are inevitable and can significantly impede accurate predictions of patient outcome. To address this challenge, we propose an \textbf{A}ttention-based Learning Framework with Dynamic \textbf{C}alibration and Augmentation for \textbf{T}ime series Noisy \textbf{L}abel \textbf{L}earning (ACTLL). This framework leverages a two-component Beta mixture model to identify the certain and uncertain sets of instances based on the fitness distribution of each class, and it captures global temporal dynamics while dynamically calibrating labels from the uncertain set or augmenting confident instances from the certain set. Experimental results on large-scale EHR datasets eICU and MIMIC-IV-ED, and several benchmark datasets from the UCR and UEA repositories, demonstrate that our model ACTLL has achieved state-of-the-art performance, especially under high noise levels.

\keywords{Time Series Classification \and Label-Noise Learning \and Medical Data Mining}

\end{abstract}

\section{Introduction}
With the increasing digitalisation of data, Electronic Health Records (EHR), which contain a wealth of patient information and statistics, have become central to medical data analysis. 
% However, the use of EHR raises concerns related to data privacy and security, necessitating more sophisticated machine-learning models for medical applications. 
EHR data are particularly susceptible to various types of errors, such as input mistakes, data discrepancies, system malfunctions, and inaccuracies in diagnostic tests \cite{Kim2017ProblemsWH}.
EHR data are usually composed of medical time series, which can encompass various forms of sequential patient-related data, such as patient visit records, vital signs monitoring, lab test results over time, etc. These data types are crucial for tracking patient progress, diagnosing trends, and predicting health outcomes. However, working with time series poses unique challenges. The data is often irregular, with inconsistent or missing timestamps due to varying patient visits or testing schedules. Additionally, medical time series data can be sparse, especially when critical events occur infrequently or when sensors or monitoring devices fail to capture continuous information. Another significant challenge is label noise, which arises due to inaccurate or incomplete annotations, misreported events, or human errors in data recording. Addressing these challenges is essential to building accurate predictive models and making reliable healthcare decisions.

While most previous label noise learning models have focused on image data within the Computer Vision (CV) domain, there remains a notable gap in addressing label noise in time series data, particularly in medical time series settings where clean annotations are severely limited.
% For example, the state-of-the-art (SOTA) model CAMELOT \cite{pmlr-v162-aguiar22a}, designed for classifying patient outcomes based on EHR, fails to produce reliable results under noisy conditions. 
In recent years, deep-learning-based robust time series classification methods such as SREA \cite{10.1007/978-3-030-86486-6_29}, which is based on self-label correction, and CTW \cite{Ma2023CTWCT}, that uses confidence-based selective data augmentation, have been developed to improve model accuracy under noisy conditions. However, SREA’s performance deteriorates rapidly with increasing noise levels, while CTW’s performance declines when the original datasets already contain sufficient high-quality instances with low noise according to empirical observation. Consequently, building reliable and trustworthy deep learning models to address these adverse phenomena, particularly for medical applications, has become a critical focus for machine learning researchers.

% which refers to a sequence of data points collected or recorded at successive time intervals, often with equal spacing between each observation. In healthcare, time series 
% these methods struggle to demonstrate their effectiveness on large-scale temporal and EHR datasets. 

To address the issue of sparsity, irregularity and label noise in large-scale EHR on patient outcome classification, we propose an \textbf{A}ttention-based Learning Framework with Dynamic \textbf{C}alibration and Augmentation for \textbf{T}ime series Noisy \textbf{L}abel \textbf{L}earning (ACTLL) tailored for EHR-related time series datasets. Our model demonstrates superior capability in capturing temporal feature representations from both local and global perspectives. Inspired by \cite{conf/icml/ArazoOAOM19}, we employ a two-component Beta Mixture Model(BMM) to divide the instances into three subsets - certain instances, uncertain instances, and hard examples for each class. Categorising these instances enables the application of distinct training strategies, thereby enhancing overall model performance. Through loss optimisation, our model consistently achieves comparable or superior performance on both large-scale EHR datasets and benchmark datasets. Our main \emph{contributions} are summarised as follows:
\begin{itemize}
    \item We design a Local-Global encoder based on Convolutional layers plus Multi-head Attention layer that effectively captures the irregular and sparse nature of medical time series data, enabling the learning of dynamic temporal patterns from both local(short-term) and global(long-term) perspectives in sequential patient visit data.
    \item We propose a novel and interpretable sample selection and division strategy, using a two-component mixture model for each class to differentiate between certain, uncertain, and hard samples rather than clean and noisy samples only.
    \item We propose a robust training strategy for dealing with label noise, by dynamically augmenting certain instances or correcting uncertain instances guided by sample selection and division.
    \item We propose a novel framework called Attention-based learning framework with dynamic Calibration and augmentation for Time series Noisy Label Learning (ACTLL). Experimental results demonstrate ACTLL's effectiveness across various EHR and benchmark datasets.
\end{itemize}

\section{Related Work}
\subsection{Label Noise Learning}
In this work, we mainly address the issue of label noise, since it is the most significant concern of the EHR learning model. Learning with noisy labels is a prominent topic in the current machine learning community, %\cite{DBLP:journals/corr/abs-2007-08199}
particularly because deep neural networks (DNNs) are prone to overfitting to noisy data. DNNs typically learn from clean samples first, but eventually overfit to noisy samples, resulting in degraded model performance \cite{Arpit2017ACL}. Label noise learning methods can be broadly categorised into two main approaches: loss correction methods and sample selection methods. 
Sample selection methods involve designing specific criteria to distinguish between clean and noisy samples. One common strategy is the small-loss criterion, which is based on the observation that noisy samples typically have higher loss values, while clean samples tend to exhibit lower loss values \cite{Chen2019UnderstandingAU}. Several extensions have been developed using this principle, such as Co-teaching \cite{Han2018CoteachingRT}, which trains two networks simultaneously, with each network being updated using the clean data selected by the other. Similarly, SIGUA \cite{Han2018SIGUAFM} utilises gradient descent on high-quality data and gradient ascent on low-quality data, which also requires knowledge of the true noise ratio. However, knowing the true noise rate is often impractical, limiting the applicability of these methods to real-world datasets. Inspired by the small-loss criterion, MixUp-BMM \cite{conf/icml/ArazoOAOM19} models a two-component Beta Mixture Model on the standard cross-entropy loss distribution and incorporates a mixup strategy to mitigate the effects of label noise. Nevertheless, neither method combines augmentation and loss correction for a robust training strategy dealing with label noise issues.

\subsection{Time Series Label Noise Learning}
Noisy label learning for time series data has recently drawn increasing attention from researchers. SREA \cite{10.1007/978-3-030-86486-6_29} is a multi-task deep learning model designed to iteratively calibrate label noise in a self-supervised manner using clustering results, label predictions, and original labels, and has demonstrated strong performance under complex industrial conditions. Similarly, CTW \cite{Ma2023CTWCT} applies time-warping augmentation to confident instances based on the normalised loss distribution of each class. However, neither method addresses the challenges of classifying large-scale medical time series, which often involves low-frequency patient visits and requires modeling temporal patterns across patient visit sequences.

\section{Methodology}
\begin{figure}[ht]% Adjust this value to move the figure left or right
    \includegraphics[width=\textwidth]{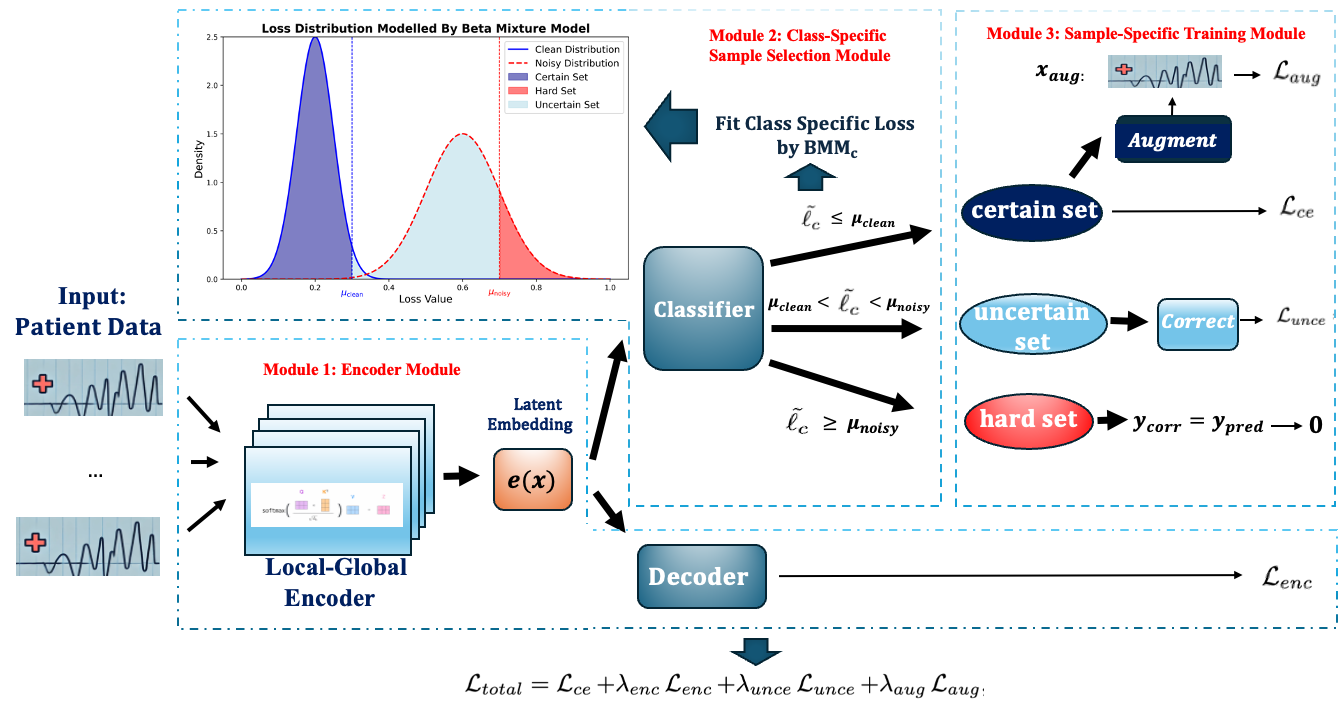}
    \caption{ACTLL Architecture}
    \label{fig:ACTLL}
\end{figure}
\subsection{Problem Statement}
The learning task of time series with noisy labels is formally defined as follows. Assuming there is a noiseless dataset $\mathcal{D} = \{(x_i, y_i)\}_{i=1}^{N}$, where $x_i$ represents the input data of instance $i$, $y_i$ represents the true class label of instance $i$ with $C$ possible values, and $N$ is the number of instances. The input data $x_i$ is the time series, where $x_i \in \mathbb{R}^{T \times F}$, $T$ is the number of time steps, and $F$ is the feature dimension. In practice, datasets often contain substantial label noise, leading to a noisy dataset defined as $\tilde{D} = \{(x_i, \tilde{y}_i)\}_{i=1}^{N}$, where $\tilde{y}_i$ is the noisy label observed, corresponding to $D$. Such label noise can be categorised as Symmetric Class-Conditionally Dependent Noise (S-CCN) or Instance-Dependent Noise (IDN), where the detailed definitions of them follow those described in \cite{10.1007/978-3-030-86486-6_29}. Our objective is to design a robust model $f^*$ that minimises the empirical risk of the model with respect to the latent true label $y_i$, rather than the noisy label $\tilde{y}_i$.
\subsection{Overview of ACTLL}
We propose a model ACTLL as illustrated in Figure \ref{fig:ACTLL}. There are three main components.
\begin{itemize}
    \item The \textbf{Local-Global Encoder} module aims to capture local(short-term) and global(long-term) dependency over sequential data and produce data representation samples during the warmup phase, as well as samples that ignore from the Sample Specific Training Module.
    \item The \textbf{Class-Specific Sample Selection} module fits a two-component Beta Mixture Model in a per-class manner to divide samples into three categories which are certain set, uncertain set, and hard set.
    \item The \textbf{Sample-Specific Training} module leverages distinct training strategies in terms of the set that instances belong to.
\end{itemize}
% The model first employs a multi-head attention-based encoder to learn sparse and temporal feature representations from sequential patient visit data, producing a low-dimensional latent embedding $e(x)$. The remaining model fitting process consists of two components: a decoder and a classifier. The decoder is designed to capture the feature representation of hard set samples, improving feature learning without the interference of label noise.

% Next, the latent embedding, $e(x)$, is fed into the classifier, which is a multi-layer perceptron trained using cross-entropy loss. The classifier’s loss is subsequently modeled using a two-component mixture model.

% The resulting loss distribution of samples is then fed into the class-specific sample selection module. The two-component Beta Mixture Model acts as the primary unsupervised model to fit the standard cross-entropy loss distribution, yielding two mixture means. We divide samples based on means of Beta Mixture Model. For the certain set, we apply time-warping augmentation throughout the training process to capture the dynamic temporal patterns in the EHR dataset. For the uncertain set, we correct label noise by leveraging both the original target and model predictions at the correction stage. This progressive learning approach improves model performance by allowing adaptive learning from all sets of data. Finally, the losses from the augmentation, certain set, and uncertain set are combined to form the final loss function.

\subsection{Local-Global Encoder}
Convolutional Neural Networks (CNNs) have demonstrated their effectiveness in handling image data. 
% However, while CNNs excel at extracting local features from images, they struggle to capture temporal information in time series data, particularly for long-range sequences. 
In contrast, attention-based models outperform Recurrent Neural Networks (RNNs) based models such as GRU or LSTM by effectively capturing both short-term fluctuations and long-term trends through attention mechanisms that focus on the critical parts of a sequence. 
% Moreover, attention-based models address the vanishing gradient issue commonly encountered in RNNs, while also enabling parallel processing, which improves both model capacity and computational efficiency.
Additionally, we introduce a multi-head attention layer, which can handle sequences of varying lengths and is robust to missing values, a common occurrence in irregular time series within EHR datasets.
\vspace{-1em}
\subsubsection{CNN Encoder.} Denoted as $\text{ConvEncoder}$, is responsible for extracting high-level features from the input time series data. A series of convolutional layers is applied to capture local dependencies and patterns across the input dimensions: $\mathbf{X} \rightarrow \text{ConvEncoder}(\mathbf{X}) \rightarrow \mathbf{H},$
where $\mathbf{X}$ represents the input time series, and $\mathbf{H}$ represents the feature map extracted by the CNN. 
\vspace{-1em}
\subsubsection{Attention Encoder.} Multi-head self-attention is applied to capture long-range dependencies in the time series. The attention mechanism computes attention scores for each time step, allowing the model to focus on different sequence parts. Defined as: $\text{Attention}(\mathbf{Q}, \mathbf{K}, \mathbf{V}) = \text{softmax}\left( \frac{\mathbf{Q} \mathbf{K}^\top}{\sqrt{d_k}} \right) \mathbf{V}$,
where $\mathbf{Q}$, $\mathbf{K}$, and $\mathbf{V}$ are the query, key, and value matrices derived from the CNN-encoded features $\mathbf{H}$, and $d_k$ is the dimension of the key. Multi-head attention is then applied, where each head learns different relationships within the sequence 
% $\text{MultiHead}(\mathbf{Q}, \mathbf{K}, \mathbf{V}) = \text{Concat}(\text{head}_1, \dots, \text{head}_h)\mathbf{W}_O$. 
Each attention head focuses on different aspects of the time series, capturing diverse temporal patterns and relationships. To improve gradient flow and stabilise the learning process, a residual connection is added, followed by layer normalisation.
% $\mathbf{H}_{\text{attn}} = \text{LayerNorm}(\mathbf{H} + \text{Attention}(\mathbf{H}, \mathbf{H}, \mathbf{H}))$.
This structure enables the model to effectively capture both short-term and long-term dependencies in the time series. We train the model to learn feature representations concurrently with other tasks by using a Mean Squared Error (MSE) loss term $ \mathcal{L}_{enc}$ for the encoder and decoder:
\begin{equation}
\label{eq:enc}
    \Loss_{enc} = \frac{1}{N}_{} \sum_{i=1}^{N} ||x_i - \hat{x_i}||_2^2,
\end{equation}
where $x_i$ is feature for instance i, $N$ is total number of instances, $||.||_2$ is $l_2$ norm.

\subsection{Class-Specific Sample Selection}
Classifying a sample as clean or noisy is crucial for optimising model performance. However, loss distributions can vary significantly across different classes, making it difficult to identify certain examples using a fixed threshold, such as the mean of the overall loss distribution. Therefore, we propose creating and fitting a two-component Beta Mixture Model $\text{BMM}_{c}$ for each class, where each component of the $\text{BMM}_{c}$ represents either the clean or noisy distribution for that specific class.
First, we normalise the data loss distribution by $z$-score normalization for each class $l_c$ to ensure that the loss values fall within the range of 0 to 1 as $\tilde{l}_c$:
\begin{equation}
\tilde{\loss_{c}} = \frac{\loss_c - \mu_c}{\sigma_c}, \quad \forall c \in \{1, \ldots, C\},
\end{equation}
where $C$ is the total number of classes in the dataset, and $c$ represents a particular class within $C$, $\mu_c$, and $\sigma_c$ represent the mean and standard deviation of loss distribution of class $c$. The probability distribution of normalised Class-Specific loss: $p(\tilde{\loss_{c}})$ can be defined as:
\begin{equation}
p(\tilde{\loss_{c}}) = \sum_{j} \theta_{j} p(\tilde{\loss_{c}}|j),p(\tilde{\loss_{c}}|j) \sim \text{Beta}(\alpha_{j}, \beta_{j}), j \in \{\text{clean}, \text{noisy}\}, \forall c \in \{1, \ldots, C\},
\end{equation}
where $\theta_j$ denotes the weight of mixture models. 
We assume that each mixture of the standardised loss distribution follows a Beta distribution, as the Beta distribution is less prone to skewing toward zero \cite{conf/icml/ArazoOAOM19}.
Using the Expectation-Maximisation (EM) algorithm \cite{conf/icml/ArazoOAOM19}, we can estimate the parameters $\hat{\alpha_j}$ and $\hat{\beta_j}$. 
To identify certain and uncertain samples, a fixed threshold may overlook confident samples due to the variability in loss distributions across different classes. Therefore, we propose using the mean of each class distribution from each mixture as the selection criterion. We consider the mixture with the smaller mean as the clean(certain) mixture formulating $\mathcal{D}_{\text{ce}}$ and the mixture with the larger mean as the noisy(hard) mixture formulating $\mathcal{D}_{\text{hard}}$, in accordance with the small-loss criterion, where we assume $\mu_{\text{clean}} \leq \mu_{\text{noisy}}$ \cite{lu2022selc}. Additionally, we regard samples with loss value between the two sets as uncertain set: $\mathcal{D}_{\text{unce}}$.
\begin{equation}
\label{eq:Dce}
    \mathcal{D}_{\text{ce}} = \{(x_{ic}, \tilde{y_{ic}}) | \tilde{\loss_{c}} \leq \mu_{\text{clean}}, \forall c \in \{1..,C\}, \forall i \in \{1,..,N\}\},
\end{equation}
\begin{equation}
\label{eq:Dunce}
    \mathcal{D}_{\text{unce}} = \{(x_{ic}, \tilde{y_{ic}}) |  \mu_{\text{clean}} < \tilde{\loss_{c}} < \mu_{\text{noisy}}, \forall i \in \{1,..,C\}, \forall i \in \{1,..,N\}\},
\end{equation}
\begin{equation}
\label{eq:Dhard}
    \mathcal{D}_{\text{hard}} = \{(x_{ic}, \tilde{y_{ic}}) | \tilde{\loss_{c}} \geq \mu_{\text{noisy}}, \forall i \in \{1,..,C\},\forall i \in \{1,..,N\}\}.
\end{equation}
This refinement enables tailored strategies for each subset, which helps mitigate the adverse impact of severe noise on the learning process.
\subsection{Sample-Specific Training Strategy}
Focusing solely on confident instances may result in the loss of valuable label information from other cases. Therefore, we propose distinct training tactics for distinct sets of data instances.
\vspace{-1em}
\subsubsection{Certain Set.}
For the certain set, the loss for this set $\mathcal{L}_{ce}$ can be defined as:
\begin{equation}
\label{eq:ce}
\Loss_{ce} = \sum_{D_{ce}} \loss(x_i, \tilde{y_i}),
\end{equation}
where $\loss(.,.)$ is the standard cross-entropy loss.
In addition to learning from the certain set, we follow the approach of CTW \cite{Ma2023CTWCT} to gradually increase the number of confident examples through Time Warping augmentation denoted as $\mathcal{D}_{\text{aug}}$, thereby ensuring the model captures temporal information from clean labels before calibrating noisy labels.
\begin{equation}
\label{eq:Daug}
    \mathcal{D}_{aug} = \{(\text{TimeWarp}(x_i),\tilde{y_i})|\forall (x_i, \tilde{y_i}) \in \mathcal{D}_{ce}\},
\end{equation}
\begin{equation}
\label{eq:aug}
    \Loss_{aug} = \sum_{\mathcal{D}_{aug}} \loss(x_i,\tilde{y_i}).
\end{equation}
\vspace{-1em}
\subsubsection{Uncertain Set.} 
For uncertain set, the main idea is to combine model predictions with observed labels, using the probability of a sample being clean or noisy as derived from the previously fitted BMM. For each sample in the uncertain set, we dynamically balance between the model’s predictions and the original noisy labels based on their respective probabilities of being clean or noisy via correction probability $h_{corr}$.
% \begin{equation}
%     h_{\text{corr}} = p(\text{noisy}|c)\tilde{y} + p(\text{clean}|c)y_{\text{pred}},
% \end{equation}
\begin{equation}
    \label{eq:corrprob}
    h_{\text{corr}} = p(\text{clean}|c)\tilde{y} + p(\text{noisy}|c)y_{\text{pred}},
\end{equation}
% \begin{equation}
% \label{eq:corrlabel}
%     y_{\text{corr}} = \argmax (h_\text{corr}),
% \end{equation} 
where $p$ denotes the softmax prediction probability, $y_{\text{pred}}$ represents the model prediction, and $\tilde{y}$ denotes the original noisy label. The term $p(\text{clean}|c)$ represents the probability that a sample is clean, while $p(\text{noisy}|c)$ captures the probability that a sample is noisy.

Eq. \ref{eq:corrprob} outlines a strategy for correcting labels in datasets with noisy annotations. In this equation, $h_{\text{corr}}$ represents the corrected label, which is calculated as a weighted combination of the original noisy label $\tilde{y}$ and the model’s predicted label $y_{\text{pred}}$. The weights, $p(\text{clean}|c)$ and $p(\text{noisy}|c)$, represent the probabilities that a sample is clean or noisy, respectively, given its class $c$. If $p(\text{clean}|c)$ is high, the model believes the original label is reliable, so $h_{\text{corr}}$ will be closer to $\tilde{y}$. Conversely, if $p(\text{noisy}|c)$ is high, indicating label corruption, the corrected label will rely more on $y_{\text{pred}}$, the model’s prediction. This formulation is particularly effective in handling samples within the ambiguous region between clean and noisy labels, allowing the model to dynamically adjust the influence of clean and predicted labels during training. Using the soft output (i.e., the probability of $h_{\text{corr}}$ before applying $\argmax$), the updated loss function for the uncertain set $\mathcal{L}_{unce}$ can be written as follows:
\begin{equation}
\label{eq:unce}
\Loss_{unce} = -\frac{1}{N} \sum_{\mathcal{D}{unce}} h_{\text{corr}}^T \log(h_{\text{pred}}),
\end{equation}
where $h_{\text{pred}}$ is the prediction probability for $y_{\text{pred}}$. Note that we use the mean instead of the sum to minimise the destabilising impact on confident learning when correction begins. As the corresponding $\lambda_{\text{unce}}$ gradually increases, the model progressively learns latent information from the uncertain set.
\vspace{-1em}
\subsubsection{Hard Set.}Using a stricter selection criterion, hard examples $\mathcal{L}_{hard}$ are considered highly unlikely to be clean. We treat their model predictions as true labels, which results in zero loss for hard set examples, i.e., $\Loss_{hard} = 0$. These examples are then used to learn their temporal feature representations via the Local-Global encoder. This strategy also mitigates the risk of the model learning incorrect label patterns, thereby preventing a deterioration in model performance.
\subsection{Training Dynamic}
We begin our model training with proper initialisation and a warmup phase using the following warmup loss function $\mathcal{L}_{warmup}$:
\begin{equation}
    \label{eq:warmup}
    \Loss_{warmup} = \sum_{\mathcal{D}} \loss(x_i, \tilde{y_i}) + \Loss_{enc},
\end{equation}
where $\mathcal{D}$ represents the entire training dataset. 
% This warmup phase facilitates a smoother transition into model learning and results in improved performance. 
Neural networks tend to memorise and learn clean labels at the beginning of training due to the simpler patterns of clean labels \cite{Arpit2017ACL}. Therefore, to maximise the learning of clean label information, we set the start of the correction phase $T_{corr}$ at 200 epochs, with correction continuing until the end of training.
% \begin{figure}[h]
%     \centering
%     \includegraphics[scale=0.45]{figures/lambda_unce.png} 
%     \caption{Value of $\lambda_{corr}$ from the time when reaching correction time until the end of the training, with $\epsilon = 1$, $n$ is a current epoch, $T_{corr}=200, T=10, k=0.1$}
%     \label{fig:lambda_unce}
% \end{figure}
When correction starts, since the model is stabilised, we gradually increase the coefficient $\lambda_{unce}$ by the following update formula:
\begin{equation}
    \lambda_{\text{unce}} = \frac{\epsilon_{\text{maxcorr}}}{1 + e^{-k \cdot ((n - T_{\text{corr}}) - T)}},
\end{equation}
where $n$ is the current epoch, $T_{\text{corr}}$ is the starting epoch for correction, $k = 0.1$ controls the steepness of the curve, which determines the rate at which the correction coefficient increases and $T = 10$ is the temperature scaling factor that controls the weight given to corrections in the early stages of scaling. The parameter $\epsilon_{\text{maxcorr}} = 1$ represents the upper bound for the correction coefficient, ensuring that $\lambda_{\text{unce}}$ is limited to the range [0, 1]. 
% We further justify our choice of $T_{\text{corr}} = 200$ in the supplementary material.
Eq. \ref{eq:trainloss} defines the final loss function used for optimisation during the correction stage. Unless otherwise specified, we set $\lambda_{\text{enc}} = 1$ and $\lambda_{\text{aug}} = 1$ for the remaining experiments.
\begin{equation}
\label{eq:trainloss}
        \Loss_{total} = \Loss_{ce} + \lambda_{enc} \Loss_{enc} + \lambda_{unce} \Loss_{unce} + \lambda_{aug} \Loss_{aug},
\end{equation}
The overall algorithm is illustrated in Algorithm \ref{alg:alg1}. 
\begin{algorithm}[h]
\caption{ACTLL: Attention-based Learning Framework with Dynamic Augmentation and Calibration for Time Series Noisy Label Learning}
\label{alg:alg1}
\begin{algorithmic}[1]
    \State \textbf{Input:} Dataset $\mathcal{D}$, learning rate $\eta$, max epochs $N$, warmup time $t_{warmup}$, correction start time $t_{corr}$
    \State \textbf{Output:} Trained Model $M$
    \State Initialise model $M$
    \For{epoch $i = 1$ to $N$}
        \If{ $i \leq t_{warmup}$}
            \State Update $\Loss_{warmup}$ using Equation (\ref{eq:warmup})
        \ElsIf{ $i \leq t_{corr}$}
            \For{class $c=1$ to $C$}
                \State Fit Beta Mixture Model $\text{BMM}_{c}$, select $\mathcal{D}_{ce}$ and create $\mathcal{D}_{aug}$
            \EndFor
            \State Update $\Loss_{total}$ using Equations (\ref{eq:enc}), (\ref{eq:ce}), (\ref{eq:aug}), (\ref{eq:trainloss})
        \Else
            \For{class $c=1$ to $C$}
                \State Fit $\text{BMM}_{c}$, select $\mathcal{D}_{ce}$, $\mathcal{D}_{unce}$, $\mathcal{D}_{hard}$, and create $\mathcal{D}_{aug}$
            \EndFor
            \State Calibrate labels from $\mathcal{D}_{unce}$ and update $\Loss_{total}$ using relevant losses
        \EndIf
    \EndFor
    \State \textbf{return} Trained Model $M$
\end{algorithmic}
\end{algorithm}
\vspace{-1em}
\section{Experiments}
\subsection{Experimental Setup}
\subsubsection{Dataset.} We conducted experiments on 13 benchmark datasets from the UCR and UEA repositories, which are widely used in evaluating state-of-the-art models, and two real-world EHR datasets MIMIC-IV-ED \cite{Johnson2021MIMIC} and eICU Collaborative Research Database \cite{Pollard2018eICU}.
\subsubsection{Baselines.}\vspace{-1em}The baseline models include: \textbf{Vanilla}: A CNN model trained using the cross-entropy loss function. \textbf{MixUp-BMM}\cite{conf/icml/ArazoOAOM19}: loss correction method based on the posterior probabilities from a two-component mixture model and the MixUp augmentation technique., \textbf{SIGUA}\cite{Han2018SIGUAFM}: Adaptive gradient descent based on the quality of different data samples, \textbf{Co-teaching} \cite{Han2018CoteachingRT}: A method employing two peer networks that select clean samples for each other. \textbf{DivideMix}\cite{li2020dividemix}: A semi-supervised robust method for image classification. \textbf{SREA}\cite{10.1007/978-3-030-86486-6_29}: A self-supervised label calibration method designed for time series learning with noisy labels. \textbf{CTW}\cite{Ma2023CTWCT}: A confidence-based Time Warping augmentation technique applied per class for time series learning with noisy labels, as current SOTA. 
\subsubsection{Implementation Details.} \vspace{-1.5em} All experiments are conducted on the High Performance Computing infrastructure, using PyTorch 1.11.0 and an NVIDIA A100 GPU. To ensure a fair comparison, we use the same hyperparameter settings as \cite{Ma2023CTWCT} and neural network topology for all experiments and perform five-fold cross-validation. We do not employ early stopping, as we assume the absence of a clean validation set, similar to real-world scenarios. The result from the last epoch is considered the outcome for that particular run, and the average of weighted F1 scores across five runs for each dataset is recorded. 
To fully explore the potential of all baseline models, we set the max epochs, $N$, for all models to 300.
The noise ratios are defined as follows: $\tau \in \{0.1, 0.2, 0.3, 0.4, 0.5\}$ for symmetric CCN, $\tau \in \{0.3, 0.4\}$ for IDN. The implementation and appendix are available at: \textbf{\url{https://github.com/yuhao-3/ACTLL}}, with additional appendix describing data information, hyperparameter analysis and visualisation.
% For our model, we set the initial learning rate to $0.001$ and halve it after reaching 20\% of the maximum number of epochs. The batch size is defined as $\min\left(\frac{1}{10} \times \text{dataset-size}, 128\right)$. Note that for the eICU dataset, we set the batch size to 512 for faster computation. The Adam optimizer parameters are set as $\beta_1 = 0.99$ and $\beta_2 = 0.999$.
% For the loss optimization parameters, we set $\lambda_{\text{aug}} = \lambda_{\text{corr}} = \lambda_{\text{enc}} = 1$. The correction start time, $t_{\text{corr}}$, is set to 200 epochs. The warmup time, $t_{\text{warmup}}$, is set to 30 epochs if the training set size is less than 1,000 samples, 15 epochs if the training set size is less than 3,000 samples, and 10 epochs otherwise. The maximum number of epochs, $N$, is set to 300 for all experiments.
\vspace{-3em}
\subsection{Comparison Result with SOTA methods.}
According to Table \ref{table2}, ACTLL achieves the best performance overall, except at 20\% symmetric. It shows comparable performance at 10\% and 20\% symmetric noise but performs significantly better as the noise rate increases. This is expected, as ACTLL can calibrate a larger number of labels for uncertain sets when the noise rate is high, and the BMM can more effectively separate clean and noisy mixtures under higher noise conditions. 
In particular, for the performance of 2 EHR datasets, ACTLL achieves the best performance with approximately 2\% increase across almost all levels. 
However, for the few noise ratios where ACTLL does not achieve the best result, MixUp-BMM, SREA, and even SIGUA predict only a single class for the eICU dataset, resulting in an inflated weighted F1-score of 0.8595 (0) for each entry. This behavior significantly overestimates the actual performance, indicating that these models are heavily influenced by label noise and therefore cannot produce reliable results under EHR-related datasets for robust learning.
\vspace{-3em}
\begin{table}[ht]
    \centering
    \caption{Average weighted F1-score and standard deviation for baseline models under different noise conditions on 13 benchmark datasets and 2 EHR datasets. The best results are highlighted in 
    \label{table2}
    \textbf{bold}.}
    \adjustbox{width=\linewidth,center}{
        \begin{tabular}{|l||c|c|c|c|c||c|c|}
            \hline
            \multirow{2}{*}{\textbf{Methods}} & \multicolumn{5}{c||}{\textbf{Symmetric Noise (\%)}} & \multicolumn{2}{c|}{\textbf{IDN Noise (\%)}} \\ \cline{2-8}
            & \textbf{10} & \textbf{20} & \textbf{30} & \textbf{40} & \textbf{50} & \textbf{30} & \textbf{40} \\ \hline
            
            \textbf{Vanilla} & 0.793(0.030) & 0.746(0.036) & 0.674(0.045) & 0.606(0.050) & 0.482(0.048)  & 0.641(0.045) & 0.563(0.044) \\ \hline
            % \textbf{CAMELOT} & 0.768 & 0.667 & 0.543 & 0.398 & 0.793 & 0.730 & - & 0.651 & 0.646 & 0.550 \\ \hline
            \textbf{SIGUA} & 0.800(0.031) & 0.766(0.035) & 0.717(0.050) & 0.654(0.060) & 0.532(0.097) & 0.670(0.059) & 0.618(0.066) \\ \hline
            \textbf{Co-teaching} & 0.797(0.030) & 0.755(0.032) & 0.691(0.046) & 0.619(0.048) & 0.514(0.053) & 0.651(0.049) & 0.575(0.049) \\ \hline
            \textbf{Mixup-BMM} & 0.771(0.033) & 0.749(0.046) & 0.734(0.037) & 0.689(0.051) & 0.553(0.064) & 0.705(0.047) & 0.636(0.053) \\ \hline
            \textbf{Dividemix} & 0.547(0.073) & 0.579(0.075) & 0.558(0.083) & 0.534(0.074) & 0.486(0.103) & 0.557(0.082) & 0.512(0.111) \\ \hline
            \textbf{SREA} & 0.817(0.028) & 0.791(0.028) & 0.750(0.033) & 0.690(0.039) & 0.552(0.062) & 0.724(0.038) & 0.686(0.061) \\ \hline
            \textbf{CTW} & 0.829(0.021) &  \textbf{0.817(0.022)} & 0.791(0.031) & 0.741(0.047) & 0.644(0.058) & 0.764(0.043) & 0.690(0.051) \\ \hline
            \textbf{ACTLL} & \textbf{ 0.831(0.020) }& 0.815(0.020) &  \textbf{0.810(0.019)} &  \textbf{0.767(0.034)} &  \textbf{0.681(0.077)} &  \textbf{0.773(0.036)} &  \textbf{0.737(0.040)} \\ \hline
        \end{tabular}
    }
\end{table}
\vspace{-3em}
\subsection{Ablation Studies}
In this section, we empirically investigate the performance of different components of our model by ablation study. We use 30\% Symm. for encoder structure, 30\%, 50\% Symm.; 40\% Ins. for sample selection setting, 30\%, 50\% Symm. as module analysis setting as representative of each ablation study.
\subsubsection{Encoder Structure.}The ACTLL model with the Local-Global Encoder as shown by Table \ref{table5a} outperforms the Diffusion Encoder by 1\% and the original CNN Encoder used in \cite{Ma2023CTWCT} by 2\%. The Diffusion Encoder works by gradually adding small Gaussian noise during the forward process and performing a reverse process for reconstruction to capture patterns within the data.
\begin{table}[ht]
    \centering
    \caption{Average weighted F1 score under Symm. 30\% and Symm. 50\% noise levels with different \textit{Encoder} and \textit{Modules}}
    \label{table5}
    \hspace{-1.5cm}
    \begin{tabular}{cc} % Two-column table layout
        % Subtable 1 (3.a)
        \begin{adjustbox}{width=0.42\textwidth}
            \begin{subtable}{0.46\textwidth}
                \centering
                \caption{Result for \textit{Encoder Structure}} % Subcaption for 3.a
                \label{table5a} % Label for 3.a
                \begin{tabular}{|l|c|}
                    \hline
                    \multicolumn{2}{|c|}{\textbf{Encoder Structure}} \\ \hline
                    \textbf{Methods} & \textbf{Sym. 30} \\ \hline
                    \textbf{ACTLL-LG} & \textbf{0.810 $\pm$ 0.019} \\ \hline
                    \textbf{ACTLL-Diffusion} & 0.772 $\pm$ 0.030 \\ \hline
                    \textbf{ACTLL-CNN} & 0.789 $\pm$ 0.030 \\ \hline
                \end{tabular}
            \end{subtable}
        \end{adjustbox}
        &
        % Subtable 2 (3.b)
        \begin{adjustbox}{width=0.45\textwidth}
            \begin{subtable}{0.56\textwidth}
                \centering
                \caption{Result for \textit{Module Analysis}} % Subcaption for 3.b
                \label{table5b} % Label for 3.b
                \begin{tabular}{|l|c|c|}
                    \hline
                    \multicolumn{3}{|c|}{\textbf{Module Analysis Result}} \\ \hline
                    \textbf{Methods} & \textbf{Sym. 30} & \textbf{Sym. 50} \\ \hline
                    \textbf{ACTLL} & \textbf{0.810 $\pm$ 0.019} & \textbf{0.686 $\pm$ 0.061} \\ \hline
                    \textbf{ACTLL w/o Aug} & 0.798 $\pm$ 0.026 & 0.671 $\pm$ 0.060 \\ \hline
                    \textbf{ACTLL w/o Corr} & 0.795 $\pm$ 0.028 & 0.668 $\pm$ 0.053 \\ \hline
                    \textbf{ACTLL w/o Aug \& Corr} & 0.794 $\pm$ 0.026 & 0.664 $\pm$ 0.053 \\ \hline
                \end{tabular}
            \end{subtable}
        \end{adjustbox}
    \end{tabular}
\end{table}
\vspace{-1em}
\subsubsection{Sample Selection Analysis.}
As observed in Table \ref{table4}, the Class-by-Class BMM strategy achieves the best results. Under a low noise ratio, the performance difference between the three methods is negligible. However, as the noise level and complexity of the noise type increase, the performance of ACTLL improves significantly, with over a 2\% increase compared to the second-best sample selection strategy for Ins. 40 setting. 
% A detailed analysis of the sample selection strategies is provided in Supplementary Material Appendix D, where we demonstrate that the effectiveness of a sample selection strategy largely depends on the properties of the datasets.
\vspace{-2em}
\begin{table}[ht]
    \centering
    \caption{Average weighted F1 score under Symm. 30\%, 50\%; Ins. 40\% noise on all datasets with different \textit{Sample Selection Strategy}}
    \label{table4}
    \adjustbox{width=0.8\textwidth,center}{
        \begin{tabular}{|l|c|c|c|}
            \hline
            {\textbf{Methods}} & \textbf{Sym 30} &\textbf{Sym. 50} &  \textbf{Ins. 40}\\ \hline
            \textbf{ACTLL-BMM} & \textbf{0.810 $\pm$ 0.019} & \textbf{0.686 $\pm$ 0.061} & \textbf{0.737 $\pm$ 0.040}\\ \hline
            \textbf{ACTLL-GMM} & 0.809 $\pm$ 0.024 & 0.679 $\pm$ 0.060 & 0.715 $\pm$ 0.042 \\ \hline
            \textbf{ACTLL-SLoss} & 0.803 $\pm$ 0.029 & 0.655 $\pm$ 0.060 & 0.706 $\pm$ 0.048\\ \hline
        \end{tabular}
    }
\end{table}
\vspace{-2em}
\subsubsection{Augmentation and Correction Module Analysis.}
As shown in Table \ref{table5b}, ACTLL performs similarly to the model without augmentation but slightly outperforms the model without correction and the model using only certain set learning at the Symm. 30\% level. When the noise level increases to 50\%, the performance gap between the original ACTLL and the ACTLL models without augmentation, without correction, and without both modules becomes more pronounced compared to the 30\% noise level.
In particular, the performance of ACTLL without the correction module is 2\% lower than that of the original ACTLL under the Symm. 50\% noise setting, highlighting the importance of the correction module under high noise levels. Although the difference between ACTLL without augmentation and ACTLL without both modules is relatively small, incorporating both modules can improve model performance by over 1.5\%. This underscores the necessity of using both the augmentation and correction modules for optimal performance.
% \begin{table}[h]
%     \centering
%     \label{table5}
%     \caption{Average weighted F1 score under symmetric 30\%, symmetric 50\%, on all datasets w/o different \textit{Modules}}
%     \adjustbox{width=0.85\textwidth,center}{
%         \begin{tabular}{|l|c|c|}
%             \hline
%             {\textbf{Methods}} & \textbf{Sym. 30} & \textbf{Sym. 50} \\ \hline
%             \textbf{ACTLL} & \textbf{0.810 $\pm$ 0.019} & \textbf{0.686 $\pm$ 0.061}\\ \hline
%             \textbf{ACTLL w/o Aug} & 0.798 $\pm$ 0.026 & 0.671 $\pm$ 0.060\\ \hline
%             \textbf{ACTLL w/o Corr} & 0.795 $\pm$ 0.028 & 0.668 $\pm$ 0.053 \\ \hline
%             \textbf{ACTLL w/o Aug \& Corr} & 0.794 $\pm$ 0.026 &  0.664 $\pm$ 0.053\\ \hline
%         \end{tabular}
%     }
% \end{table}

\section{Conclusions}
In conclusion, we propose a novel method for EHR time series data with noisy labels, named ACTLL. Our approach performs label correction using the posterior probabilities of the Beta Mixture Model (BMM) for uncertain data instances, with the BMM model applied on a class-by-class basis. Instead of using a fixed threshold, we leverage the mean of the clean and noisy distributions to achieve better separation and higher sample selection accuracy, thereby avoiding sample selection bias. Experiments conducted on various datasets demonstrate that ACTLL outperforms other state-of-the-art methods not only on EHR datasets but also on benchmark datasets. However, our model incurs higher time complexity due to the construction of BMM for each class, which is necessary for achieving better sample selection accuracy. Additionally, improving model performance under low-noise scenarios and imbalanced scenarios, which are common in EHR datasets, remains a topic for future research.

\end{document}